\documentclass[10pt,twocolumn,letterpaper]{article}

\usepackage[]{iccv}      
\definecolor{iccvblue}{rgb}{0.21,0.49,0.74}
\usepackage[pagebackref,breaklinks,colorlinks,allcolors=iccvblue]{hyperref}
\usepackage{multirow}
\usepackage{makecell}

\usepackage{fancyvrb}

\def\paperID{13722} 
\def\confName{ICCV}
\def\confYear{2025}

\title{CI-VID: A Coherent Interleaved Text-Video Dataset
}

\author{
    \textbf{Yiming Ju}\textsuperscript{\rm 1*}, 
    \textbf{Jijin Hu}\textsuperscript{\rm 2*}, 
    \textbf{Zhengxiong Luo}\textsuperscript{\rm 1*}, 
    \textbf{Haoge Deng}\textsuperscript{\rm 2*}, 
    \textbf{hanyu Zhao}\textsuperscript{\rm 1}, 
    \textbf{Li Du}\textsuperscript{\rm 1}, \\ 
    \textbf{Chengwei Wu}\textsuperscript{\rm 1}, 
    \textbf{Donglin Hao}\textsuperscript{\rm 1}, 
    \textbf{Xinlong Wang}\textsuperscript{\rm 1 \dag}, 
    \textbf{Tengfei Pan}\textsuperscript{\rm 1 \dag} \\
    \textsuperscript{\rm 1}  Beijing Academy of Artificial Intelligence \\
    \textsuperscript{\rm 2} Beijing University of Posts and Telecommunications  \\
    \texttt{\{ymju, tfpan, wangxinlong\}@baai.ac.cn} \\
}

\begin{document}
\maketitle


\renewcommand{\thefootnote}{}
\footnotetext{\textsuperscript{*}Equal Contribution. \textsuperscript{\dag}Corresponding Author.}
\renewcommand{\thefootnote}{\arabic{footnote}}

\begin{figure*}[t]
  \centering
\includegraphics[page=1, width=1.0\linewidth]{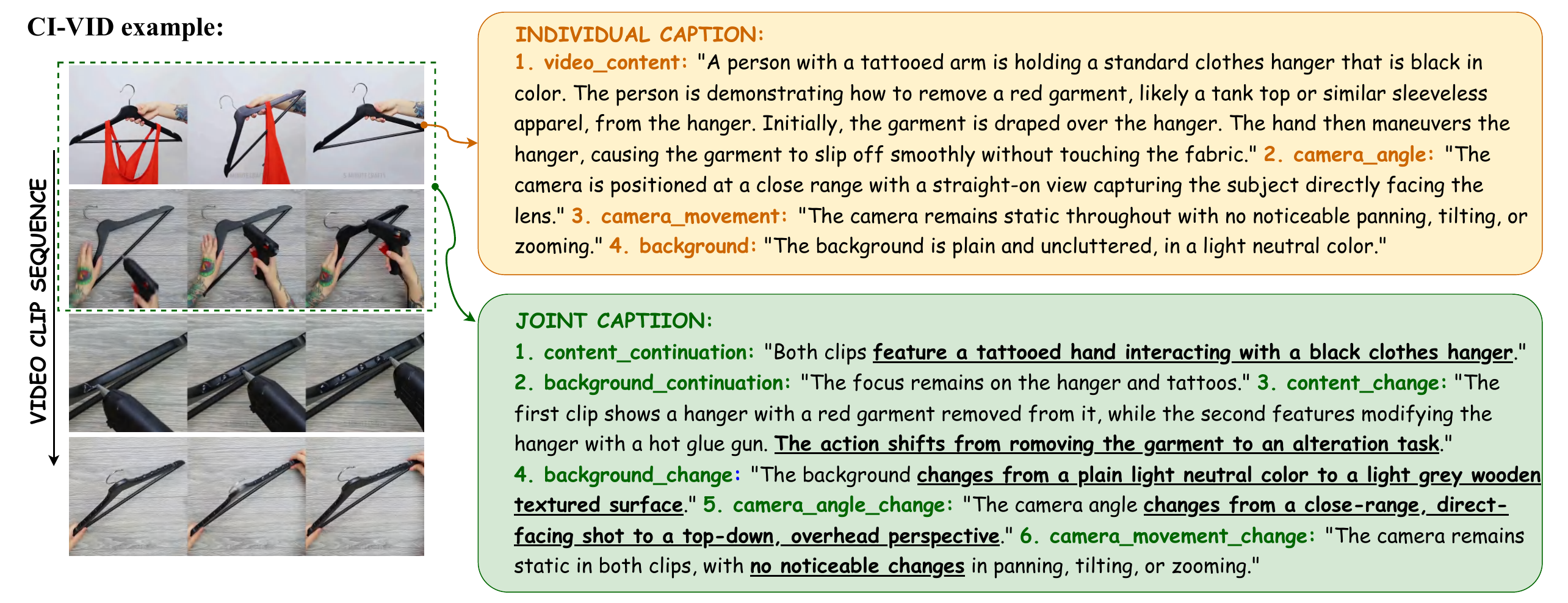}
   \caption{
An example from the CI-VID dataset. Each sample consists of a sequence of video clips, individual captions describing each clip, and joint captions capturing the continuity and change between adjacent clips.
}
   \label{sample_show}
\end{figure*}

\begin{abstract}

Text-to-video (T2V) generation has recently attracted considerable attention, resulting in the development of numerous high-quality datasets that have propelled progress in this area.
However, existing public datasets are primarily composed of isolated text-video (T-V) pairs and thus fail to support the modeling of coherent multi-clip video sequences.
To address this limitation, we introduce CI-VID, a dataset that moves beyond isolated text-to-video (T2V) generation toward text-and-video-to-video (T\&V2V) generation, enabling models to produce coherent, multi-scene video sequences.
CI-VID contains over 340,000 samples, each featuring a coherent sequence of video clips with text captions that capture both the individual content of each clip and the transitions between them, enabling visually and textually grounded generation.
To further validate the effectiveness of CI-VID, we design a comprehensive, multi-dimensional benchmark incorporating human evaluation, VLM-based assessment, and similarity-based metrics.
Experimental results demonstrate that models trained on CI-VID exhibit significant improvements in both accuracy and content consistency when generating video sequences.
This facilitates the creation of story-driven content with smooth visual transitions and strong temporal coherence, underscoring the quality and practical utility of the CI-VID dataset
We release the CI-VID dataset and the accompanying code for data construction and evaluation at:\
\href{https://github.com/ymju-BAAI/CI-VID}{https://github.com/ymju-BAAI/CI-VID}

\end{abstract}
\section{Introduction}

Recent advances in Artificial Intelligence Generated Content (AIGC) have been largely driven by growing data and compute \citep{ju2025miradata}. In the field of computer vision, the success of recent text-to-video (T2V) models, such as Sora \citep{brooks2024video}, VideoPoet \citep{kondratyuk2023videopoet}, Emu3 \citep{wang2024emu3}, CogVideoX \citep{yang2024cogvideox}, and VideoTetris \citep{tian2024videotetris}, has notably expanded the possibilities for visual content generation, enabling the automatic creation of hyper-realistic videos based on human instructions. 

Researchers have contributed many high-quality video generation datasets to advance the field, including OPENVID \citep{nan2024openvid}, InternVid \citep{wang2023internvid}, ShareGPT4Video~\citep{chen2024sharegpt4v}, and Vript~\citep{yang2024vript}, among others.
Although these datasets provide high-quality video clips paired with text captions, most consist solely of isolated text–video (T–V) pairs in a one-to-one correspondence, without modeling inter-clip relationships or temporal coherence.
This one-to-one pairing paradigm presents two main limitations:

\textbf{1. T2V models trained solely on independent T-V pairs cannot generate consistent cross-scene videos
}.
Existing datasets typically segment videos at scene boundaries and annotate each clip independently.
However, real-world videos often consist of multiple semantically connected scenes that are content-related but visually disjoint due to changes in camera angle, entities, or location.
For example, Figure~\ref{sample_show} presents a tutorial on modifying a black hanger using a glue gun, where the content is conveyed through a sequence of clips, each contributing partial information toward constructing the complete scene.
Due to the one-to-one correspondence in previous datasets, T2V models trained on them struggle to generate cross-scene video sequences with consistent characters, coherent visual style, and smooth scene transitions.

\textbf{2. Independent T–V pairs do not support training text-and-video-to-video (T\&V2V) generation models.}
In video extrapolation tasks, prior methods typically rely solely on preceding visual frames as input~\cite{mcvd}, which frequently results in repetitive generations~\citep{tats} and lacks semantic control.
To guide the extrapolated content meaningfully, textual inputs are crucial as conditioning signals.
However, existing datasets—primarily composed of isolated T–V pairs—are inherently unsuited for learning generation conditioned jointly on both visual and textual inputs.
As a result, they are inadequate for training models capable of T\&V2V generation.

These limitations render the independent T–V pair format inadequate for complex applications beyond unit-level text-to-video generation, such as storytelling, video rewriting, and other advanced video generation tasks.
To bridge this gap, we introduce CI-VID—a novel, carefully curated dataset of Coherent Interleaved Text–Video sequences.
Figure~\ref{sample_show} shows a CI-VID sample consisting of a sequence of video clips, each paired with an individual caption, as well as joint captions that describe the continuity and distinctions between adjacent clips.
As illustrated, CI-VID provides inter-clip relational information, which is not available in prior video generation datasets.
CI-VID exhibits several key characteristics:


\begin{itemize}

    \item \textbf{High-Quality Video Content.} CI-VID sources its videos from over 4,000 carefully curated YouTube channels spanning a wide range of themes. Clips are rigorously filtered based on on-screen text ratio, motion differences, and visual clarity, with fewer than 20\% retained for further processing.

    \item \textbf{Content-Relevant but Visually Diverse Sequences.} 
    Video clip sequences in CI-VID are designed to preserve visual diversity while maintaining narrative coherence.
    Consistency in style, entities, and visual details allows previous clips to serve as base input for generating subsequent ones. While variations—such as shot transitions, action changes, and new entities—enable textual descriptions to provide meaningful guidance, supporting instruction-following rather than simply replicating patterns from visual input.
    
    \item \textbf{High-Quality Text Captions.} CI-VID provides detailed and structured captions that go beyond per-clip descriptions by capturing both the continuity and distinctions between adjacent clips. These enriched annotations support video generation with joint guidance from both video and text inputs.
    
\end{itemize}

CI-VID comprises over 340,000 high-quality samples.
To assess its effectiveness and evaluate the task of coherent video sequence generation, we construct a multi-dimensional benchmark incorporating human evaluation, vision-language model (VLM)-based assessment, and similarity-based metrics.
Experimental results show that models fine-tuned on CI-VID can effectively leverage both preceding visual context and textual instructions to guide generation, significantly outperforming baselines in producing coherent, story-driven video sequences with smooth transitions and consistent content.

Our contributions are summarized as follows:

1. We identify the limitations of isolated T–V pair training data and argue that future datasets should support not only T-to-V mapping but also T\&V-to-V modeling to enable coherent and controllable video sequence generation.

2. We introduce CI-VID, a high-quality dataset for T\&V-to-V generation. CI-VID consisting of interleaved text–video sequences with coherent multi-clip videos and captions that describe both individual content and inter-clip transitions.

3. We establish a comprehensive multi-dimensional benchmark for the task of coherent video sequence generation and conduct preliminary experiments based on CI-VID.

\begin{figure*}[t]
  \centering
   \includegraphics[width=1.0\linewidth]{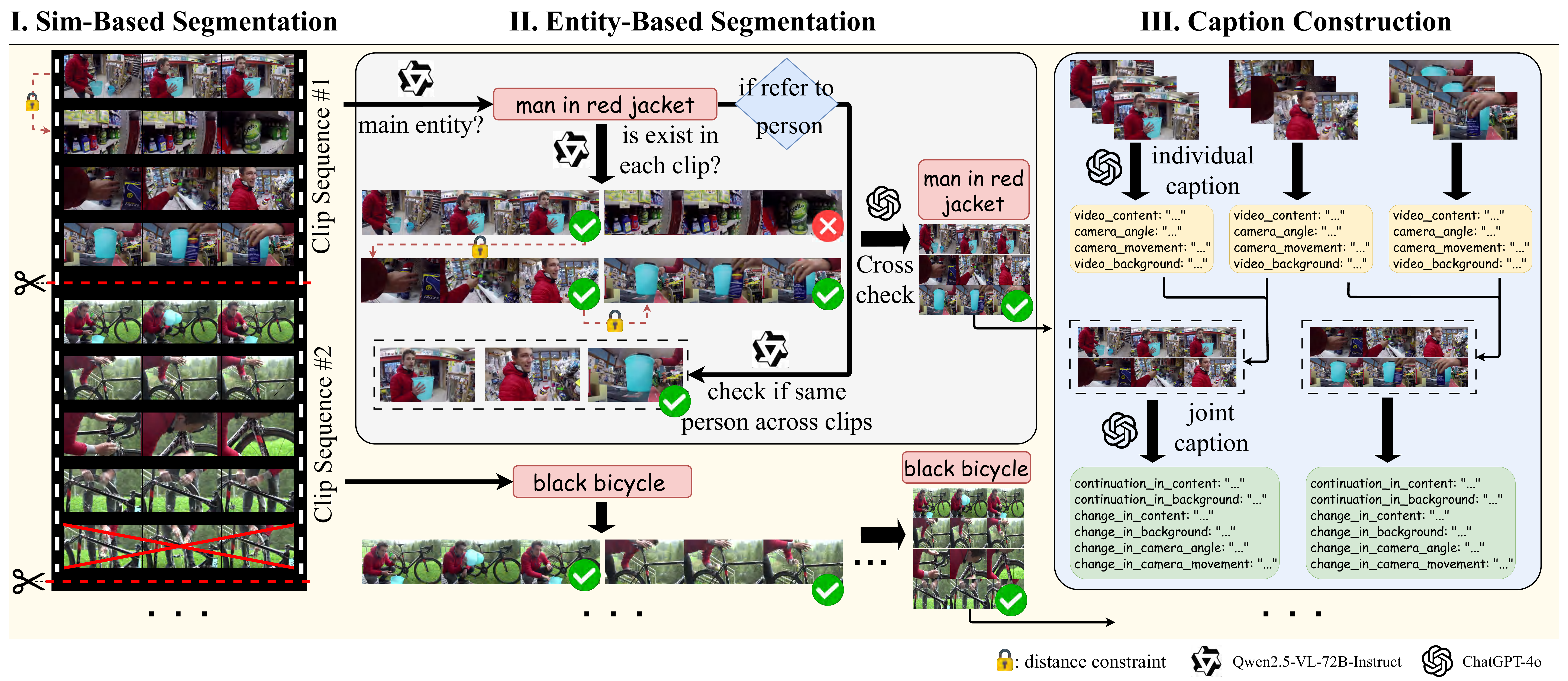}
\caption{The pipeline for constructing CI-VID samples.
The Segmentation modules construct clip sequences from the source video, while the Caption Construction module generates individual captions for single clips and joint captions for adjacent clips.}

   \label{pipeline}
\end{figure*}

\section{Related Work}
\subsection{Text-to-Video Datasets}
Recent advancements, such as Sora \cite{brooks2024video} and VideoPoet \citep{kondratyuk2023videopoet}, demonstrate the promising potential of T2V generation. Building powerful T2V models requires high-quality video-text datasets for vision-language alignment. Researchers have contributed numerous high-quality datasets to support the development of this field.
WebVid-10M \cite{bain2021frozen} uses web crawling to collect video-text pairs;
HDVILA-100M \cite{xue2022advancing} uses titles, descriptions, captions to build video-text pairs;
However, as pioneering datasets WebVid-10M and HDVILA-100M suffer from limitations such as low video resolution, watermarks, and noisy captions.
Panda-70M \cite{chen2024panda} extends HDVILA-100M by filtering video clips for scene consistency. It then employs multimodal models to generate more accurate captions;
OpenVid-1M \cite{nan2024openvid} provides high-quality, diverse video samples, addressing the shortcomings of WebVid-10M and Panda-70M;
InternVid \cite{wang2023internvid} collects web videos based on action and activity keywords. It generates frame-by-frame descriptions and summarizes them with language models to produce more informative captions;
Vript \cite{yang2024vript} focuses on generating highly detailed captions, using a multi-stage description generation process. 
It has an average of 150 words per caption, significantly longer than the under 30 words typical of previous datasets;
MiraData \cite{ju2025miradata} addresses the short video length issue in existing datasets by selecting specific channels and merging similar slices, achieving an average video length of 70 seconds—much longer than the typical 20 seconds in other datasets.
ShareGPT4Video \cite{chen2024sharegpt4v} extracts key frames from videos and applies differential caption strategy to generate temporally ordered descriptions that capture key actions.
As introduced, the data form of current T2V datasets are confined to a one-to-one text-video correspondence, without considering the connection among video clip.

\subsection{Interleaved Datasets}
The concept of interleaved data originates from the image-text domain, referring to sequences where text and images are interwoven consecutively. 
Studies such as Flamingo \cite{alayrac2022flamingo} and KOSMOS-1 \cite{huang2023language} demonstrate that models trained on interleaved datasets outperform those trained on image-description pairs, highlighting the benefits of leveraging correlations in interleaved content. However, the datasets used in Flamingo and KOSMOS-1 are not publicly available. To address this, several open-source interleaved image-text datasets have been introduced;
MMC4 \cite{zhu2023multimodal} extends the text-only C4 corpus \cite{raffel2020exploring} by incorporating images into text passages utilizing CLIP \cite{radford2021learning} features;
OBELICS \cite{laurenccon2023obelics} extracts interleaved sequences from web page content through rigorous filtering techniques;
CoMM \cite{chen2024comm} collects raw data from specific websites and employs multiple models to filter out incoherent text-image pairs.
However, interleaved datasets are primarily limited to the image-text domain and remain unexplored in the field of video generation.

\section{CI-VID Dataset}

To address the limitations of independent T-V pair training data, we constructed CI-VID, a large-scale interleaved text-video dataset. CI-VID consists of narratively coherent and thematically consistent video clip sequences, with structured and detailed captions describing each clip and the relationships between adjacent clips.
Figure \ref{pipeline} illustrates the pipeline for constructing CI-VID samples from raw videos. The overall construction process consists of three main stages: source video collection, video clip sequence construction, and caption generation. 

\subsection{Source Video Collection and Processing} 
CI-VID construction requires complete source videos rather than pre-segmented clips.
Thus, we collect raw videos from YouTube, similar to many existing public datasets \cite{xue2022advancing, chen2024panda, wang2023internvid, yang2024vript, ju2025miradata, chen2024sharegpt4v}, rather than relying on existing large-scale video datasets such as Panda-70M and HDVILA-100M.

~\\
\noindent
\textbf{Source Video Collection.} Although YouTube offers a vast range of video types, many videos suffer from low quality and do not meet the requirements for video generation. To ensure the quality of source videos, we collected videos by channel first.
Specifically, we utilized the training data of Emu3 \cite{wang2024emu3} and extracted the corresponding channels associated with these training samples.
We then manually filtered 4,068 high-quality channels from this list. Annotators assessed video quality within these channels based on factors such as resolution, color fidelity, motion strength, and watermark presence, without imposing content restrictions\footnote{The annotation team consisted of six professional annotators, each holding a bachelor's degree. They underwent training with 200 sample cases and were required to review at least three videos per channel.}.
We downloaded all public videos from selected channels and ultimately obtained 592,429 raw videos.

~\\
\noindent
\textbf{Segmentation and Filtering.} The raw videos were first segmented into clips using content-aware detection of PySceneDetect\footnote{https://github.com/Breakthrough/PySceneDetect}  with a threshold of 3, ensuring that each clip contained a single shot.
Long-duration clips were evenly split to ensure that no clip exceeded ten seconds.
Thus CI-VID includes both independent shot clips and clips derived from splitting continuous shots, which account for 35.2\% of the total clips and are specially marked.
Moreover, clips shorter than one second are filtered to ensure sufficient duration for training data.
Next, optical flow \cite{teed2020raft} is calculated to maintain adequate motion strength. 
The average flow magnitude per pixel normalized by the shorter edge, is used to filter out clips below the acceptable threshold (70).
Finally, text detection is performed using PaddleOCR\footnote{https://github.com/PaddlePaddle/PaddleOCR}, and clips with excessive text coverage (over 10\%) were discarded
Overall, these filtering processes remove over 80\% of the clips.

\subsection{Video Clip Sequence Construction}

Constructing video clip sequences is the most critical step in the building of CI-VID. Modeling T\&V2V generation requires video clips in one sequence to be sufficiently related so that the clip can serve as part of the control information for generating, such as maintaining consistency in style, characters, background, and visual details.
At the same time, clips also need to maintain enough variation to allow textual descriptions to provide meaningful guidance, such as shot transitions, action changes, the introduction of new entities, and background shifts.
These variations can train the model's ability to adhere to text instructions.
Simply extracting consecutive clips from the source video fails to meet these requirements. Thus, CI-VID employs a carefully designed pipeline for constructing clip sequences, as shown in Figure \ref{pipeline}, which includes two steps: Similarity-Based Segmentation and Entity-Based Segmentation.

\subsubsection{Similarity-Based Segmentation}

The core idea of similarity-based segmentation is to assess the correlation and variation between video clips based on embedding similarity, thereby segmenting raw video into distinct sequences.
Specifically, we define a high similarity threshold $T_{h}$ and a low similarity threshold $T_{l}$. 
If the similarity between the current clip and the previous clip falls below the $T_{l}$, it is identified as a scene transition, and a new sequence is initiated with the current clip. Conversely, if the similarity exceeds $T_{h}$, the current clip is considered to provide little variation and is ignored. Only when the similarity falls within the threshold range is the current clip added to the ongoing sequence.

To compute similarity, we extract three frames from each clip at equal intervals and concatenate them horizontally, as illustrated in Figure \ref{pipeline}. 
Then, ImageBind model \cite{girdhar2023imagebind} is used to obtain the clip’s vector representation, with cosine similarity as the similarity metric.\footnote{We found that widely used intermediate/key frame similarity was significantly less effective in detecting scene transitions than concatenating multiple frames. This may be because concatenating multiple frames provides a more comprehensive representation of the video clip and explicitly encodes temporal correspondences through spatial positioning.}
The thresholds were empirically set to $(T_{l}$, $T_{h}) = (0.6, 0.8)$, adopting a strict range to prioritize segmentation quality over sample quantity.
Filtering in the Source Video Collection process (\eg, motion detection) and the removal of highly similar clips may result in discontinuous clips within a sequence. Therefore, we enforce two distance constraints to preserve scene continuity:

\begin{itemize}
    \item The index difference between adjacent clips (clip index before filtering) must not exceed three.
    \item The time gap between adjacent clips must not exceed ten seconds.  
\end{itemize}

Nonconforming points are treated as scene transitions, and the sequences are split accordingly.
Finally, sequences containing only a single clip are discarded, yielding the initial set of segmented clip sequences from the source video.

\begin{table}[t!]
\small
\centering
\renewcommand{\arraystretch}{1.3}
\begin{tabular}{p{7.5cm}}
\toprule
\emph{In this figure, each column contains an image. Can you identify the most
common entity (objects/people/goals) among these images? Note:} \\
\emph{1) Only return the most common entity. } \\
\emph{2) The entity must be the same one. } \\ 
\emph{3) The entity must be the main entity, not the background or edge entity.} \\ 
\emph{4) The entity must appear in more than 60\% of the images. Return 'none' if there
are none.} \\ 
\emph{5) Return the entity name directly, with its characteristics. }\\ 
\emph{6) The same person is also an entity, return person‘s characteristics(hair, dress),
don't guess person‘s name.} \\
\bottomrule
\end{tabular}
\caption{Prompt for extracting the main entity.}
\label{prompt}
\end{table}

\subsubsection{Entity-Based Segmentation}

Due to the diversity and complexity of video content, it is extremely difficult to ensure content relevance through embedding similarity alone.
Thus, we propose Entity-Based Segmentation to further refine results generated by the Similarity-Based Segmentation module.
The core idea is that if a series of clips share a common entity, they are considered content-related.
As illustrated in Figure \ref{pipeline}, the segmentation process consists of four main steps:

\begin{itemize}
\item \textbf{main entity extraction}: We employ Qwen2.5-VL-72B-Instruct \cite{bai2025qwen2}, one of the most powerful visual understanding models, to extract the main entity of a clip sequence. The input consists of a 3 × $n$ grid image, where $n$ is the clip sequence length, and each row contains three frames evenly sampled from one clip. The query prompt is shown in Table \ref{prompt}. If no main entity is detected, the sequence is discarded.

\item \textbf{clip entity examination}: Each of the three frames from one clip is individually used as input with querying whether it contains the main entity. If at least one frame does, the clip is considered to pass the examination. If fewer than 70\% of the clips pass the examination, the entire sequence is discarded. Furthermore, clips that fail to pass the examination are removed from the sequence.  

\item \textbf{same-person verification}: Different individuals in a video may share similar visual features, such as clothing and hair color, leading to potential confusion\footnote{For example, if the main entity is ``a person with a black T-shirt," Clip\#1 contains Person\#A wearing one, while Clip\#2 contains Person\#B wearing another. Although both clips pass the entity examination, Person\#A and Person\#B are different individuals.}. 
To mitigate this issue, we select one frame containing the main entity from each clip and merge them into a single image. Then, we query the model to determine whether the main entities in each clip are the same person. If they are not, the sequence is discarded.  

\item \textbf{revalidation}: The previous three steps rely on Qwen2.5-VL-72B-Instruct. Finally, we perform cross-validation using GPT-4o \cite{hurst2024gpt}, one of the most powerful visual understanding models, to detect and discard erroneous sequences, further improving sequence quality.   
This step follows the same image input format as the first step, and the model is asked to verify whether the sequence and main entity meet the original requirements shown in Table \ref{prompt}. 

\end{itemize}

By incorporating entity-based segmentation, we enhance the coherence of clip sequences beyond what is achievable with similarity-based methods. This step ensures that adjacent clips not only exhibit sufficient visual distinction but also maintain strong relevance in content.

\subsection{Caption Generation} 
To support T\&V2V generation, it is essential to have not only highly correlated clip sequences but also corresponding text descriptions to help the model understand the relationships among clips.
Thus, we not only provide individual captions for each clip but also generate joint captions that capture the relationships between adjacent clips, as shown in Figure \ref{sample_show}.

We leverage the powerful visual understanding capabilities of GPT-4o for caption generation.
We found that the sequential-frame-input strategy—feeding frames into the model sequentially—produces more detailed and accurate descriptions, capturing intricate background compositions and fine-grained object features.
In contrast, the joint-frame-input strategy—combining multiple frames into a single large image—better captures overall scene relationships, such as character transitions and shifts in perspective or background.
Thus, we adopt a two-step caption generation pipeline, as illustrated in Figure \ref{pipeline}.
First, individual captions are generated using the sequential-frame input strategy to capture fine-grained details.
Then, joint captions are generated using the joint-frame input strategy to better capture scene transitions.

Specifically, for individual caption generation, we sample 4–8 frames per clip at even intervals based on its duration and construct structured captions covering four key aspects: \textit{video content}; \textit{camera angle}; \textit{camera movement}; and \textit{video background}.
For joint caption generation, we use an $x$ × 2 grid image as input, where each row contains $x$ frames sampled evenly from a clip.
The value of $x$ ranges from 3 to 5, depending on the longer duration between the two clips.
We then construct joint captions from the following perspectives:
\textit{continuation in video content};
\textit{change in video content};
\textit{continuation in video background};
\textit{change in video background};
\textit{change in camera angle}; and
\textit{change in camera movement}.\footnote{Since video content and background are usually more complex than camera angle and movement, and are neither strictly unchanged nor entirely different entirely different across clips.
Thus, joint captions describe both continuation and change aspects to accurately capture the relationships between clips.}
Finally, for a clip sequence of length $n$, we obtain $n$ individual captions and $n -1$ joint captions.

\begin{table}
  \centering
  \resizebox{\linewidth}{!}{
  \begin{tabular}{lcccc}
    \toprule
    \multirow{2}{*}{Dataset} & Clip  & Clip  & Caption  & \multirow{2}{*}{Year} \\
     &  num &  duration &  length (words)&  \\

    \midrule
    HowTo100M & 136M & 3.6s & 4.0 & 2019\\
    WebVid-10M & 10M & 18.0s & 12.0 & 2021\\
    InternVid & 234M & 13.4s & 32.5 & 2023\\
    HD-VG-130M & 130M  & 5.1s & 9.6 & 2024\\
    Panda-70M & 70M & 8.5s & 13.2 & 2024\\
    MiraData &  798K & 72.1s & 318.0 & 2024\\
    Vript & 420K  & 11.1s & 145.0 & 2024\\
    OPENVID-1M & 1M & 7.2s & 98.3 & 2024\\
    \midrule[0.2pt]
    \rowcolor[gray]{0.9}
    CI-VID$_{individual}$ & 1M & 4.7s & 218.6 & 2025\\
    \rowcolor[gray]{0.9}
    CI-VID$_{joint}$ & 717K & 9.6s & 215.8 & 2025\\

    \bottomrule
  \end{tabular}
  }
  \caption{Comparison of CI-VID statistics with existing large-scale video-text datasets.}
  \label{comparison}
\end{table}

\subsection{Analysis of CI-VID}

CI-VID comprises a total of 341,550 samples, with each sample containing an average of 3.1 video clips.
Over 98\% of the video clips have a resolution of 1080p or higher.
Table \ref{comparison} presents a comparison of clip number and caption length between CI-VID and recent video-text datasets\footnote{CI-VID$_\text{joint}$ refers to treating each pair of adjacent clips as a single unit, where the clip number indicates the total number of such units, the clip duration represents the combined duration of the two adjacent clips, and the caption corresponds to the joint caption.
}.
As shown in the table, CI-VID's clip duration is slightly shorter than recent datasets, resulting from the use of a very low PySceneDetect threshold (3) to ensure no shot transitions within a single clip\footnote{Existing datasets typically use higher thresholds, such as MiraData (26), InternVid (27), and Panda-70M (25), leading to coarser segmentation. However, shot transitions introduce rapid optical changes, often regarded as noise during training video generation models.}.
Regarding caption length, both individual and joint captions exceed 200 words on average, second only to MIRA. This reflects fine-grained descriptive granularity, highlighting CI-VID’s significant value as a video-text dataset.

\begin{figure}[t]
  \centering
   \includegraphics[width=1.0\linewidth]{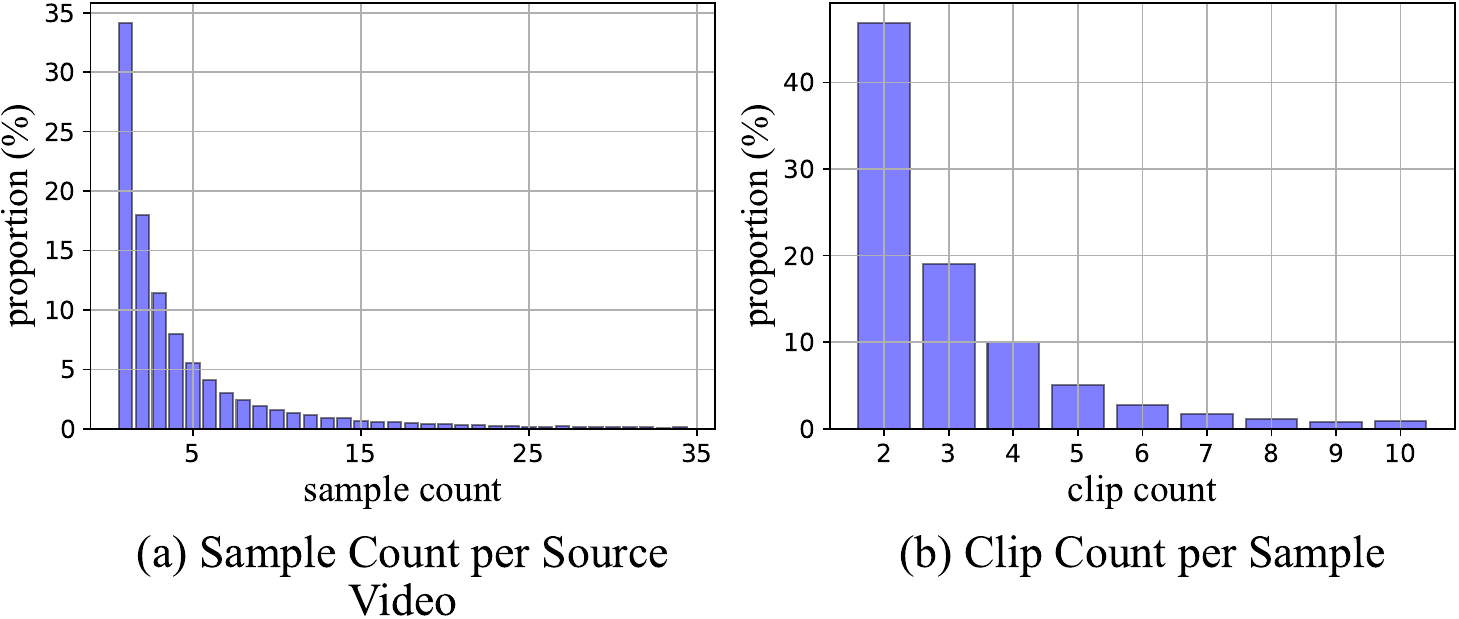}

\caption{The analysis of sample characteristics: (a) the number of samples generated per source video. (b) The distribution of clip sequence length.}   \label{ana}
\end{figure}

\begin{figure}[t!]
  \centering
   \includegraphics[width=0.82\linewidth]{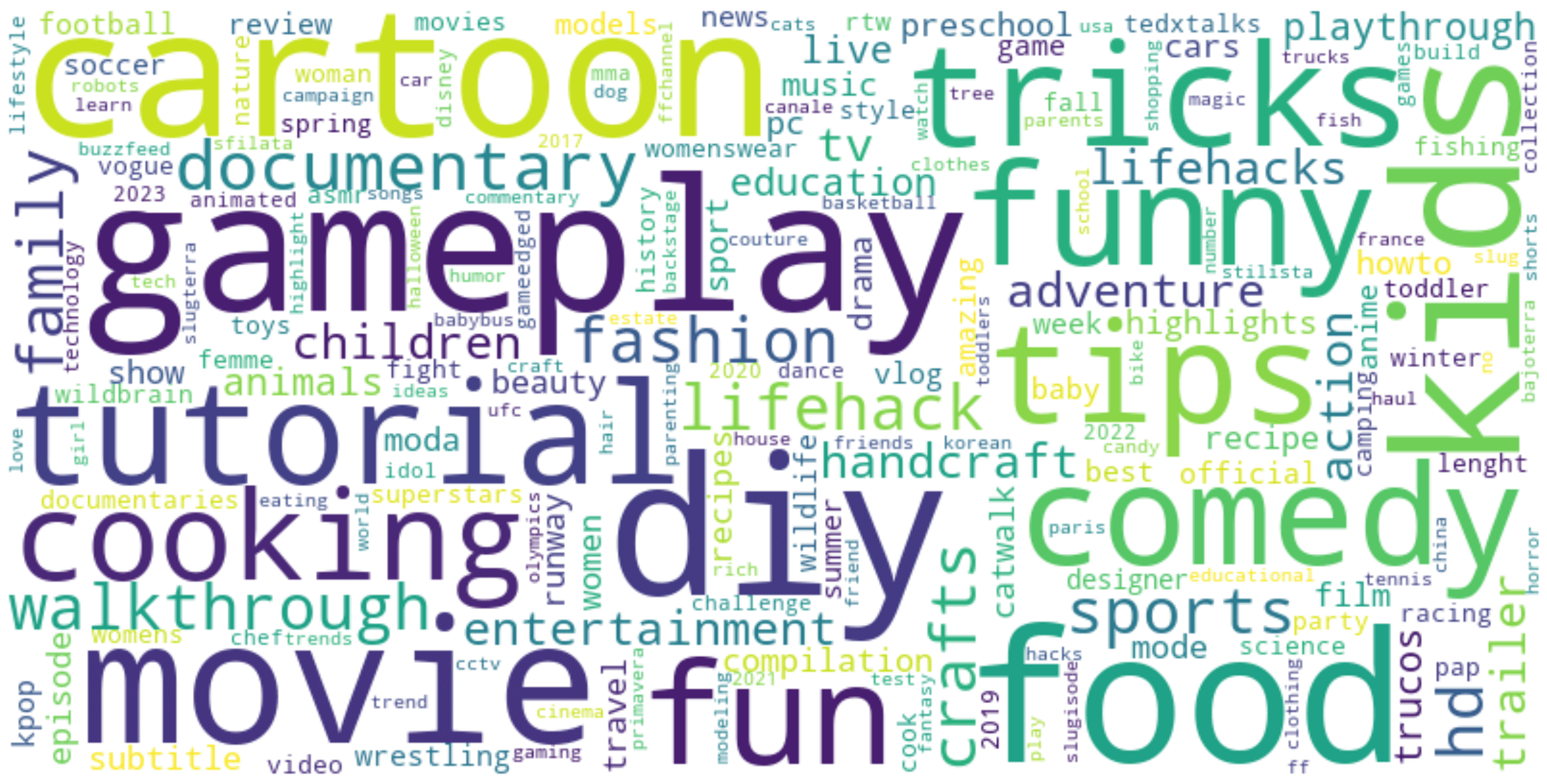}
   \caption{Word cloud of CI-VID sample themes, derived from corresponding YouTube tags.}
   \label{cloud}
\end{figure}

CI-VID samples are derived from 63,807 original videos. Figure \ref{ana}(a) shows the number of samples generated per source video, showing that most videos contribute fewer than ten samples. 
This indicates that CI-VID avoids an overrepresentation of a small set of videos, ensuring dataset neutrality.
Figure \ref{cloud} presents the word cloud of CI-VID sample themes, derived from the corresponding YouTube tags of each sample.
The visualization highlights a diverse range of video categories, including food, entertainment, education, and sports, further demonstrating the rich content diversity of the CI-VID dataset.
Figure \ref{ana}(b) presents the distribution of clip counts per sample (clip sequence length), showing that while most samples contain 2–3 clips, over 30\% (more than 100K samples) include four or more clips. This characteristic makes CI-VID valuable for both pairwise training scenarios and those requiring long input sequences.


\begin{figure*}[t!]
  \centering
   \includegraphics[width=1\linewidth]{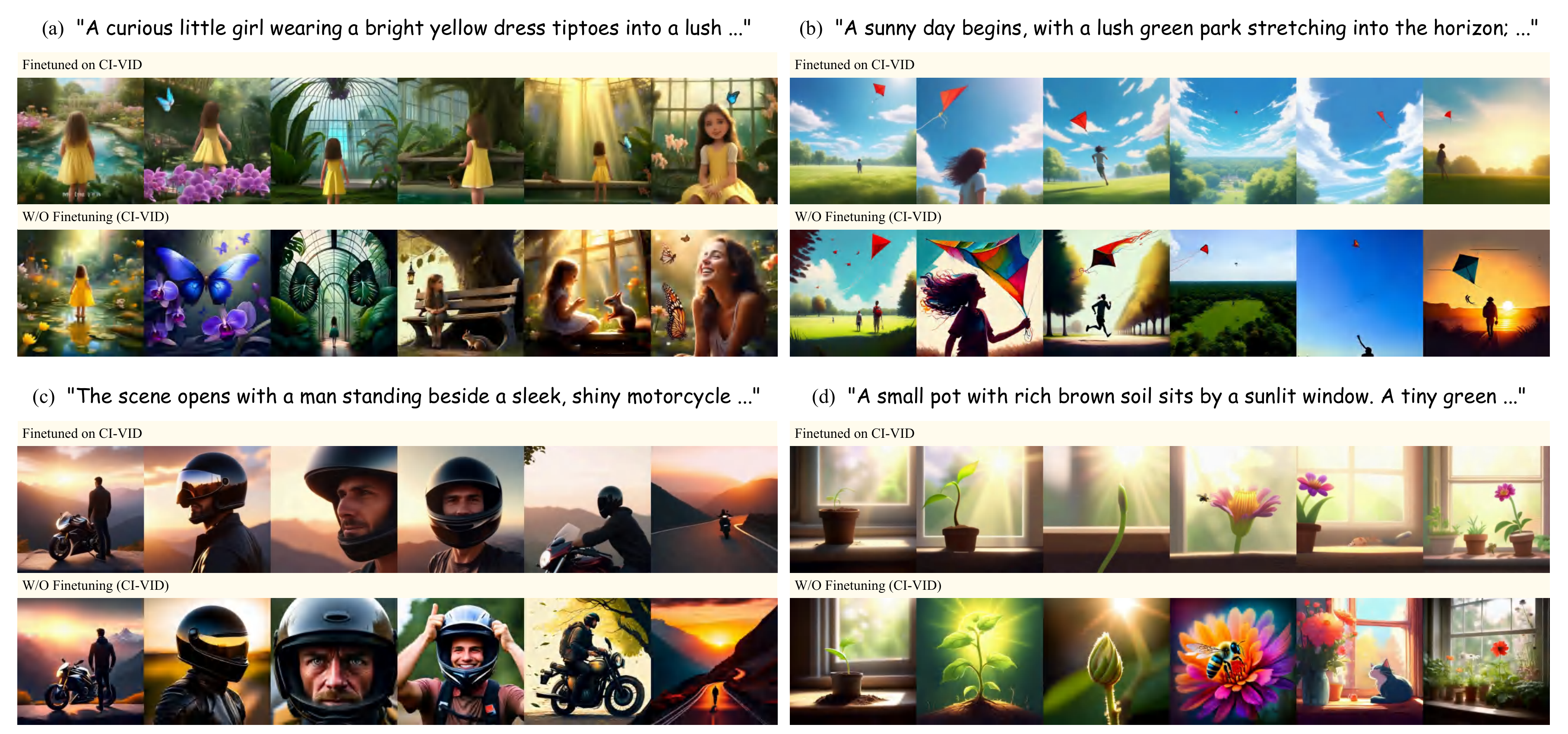}
\caption{Comparison between generated results with and without finetuning on CI-VID. Sample (a) shows the generated result from the prompt in Table~\ref{tab:prompt-example}.}
\label{fig:qualitative}
\end{figure*}






\section{Experiment}
To validate the effectiveness of CI-VID and evaluate coherent video sequence generation, we trained a small-scale video generation model and designed a comprehensive multi-dimensional benchmark to assess its performance.


\subsection{Experiment Settings} 
\textbf{Model Setting.}
We primarily follow the approach of NOVA~\cite{deng2024nova} for sequentially predicting temporal frames to process interleaved text-video data in CI-VID.
Our model comprises a temporal encoder, a spatial encoder, and a decoder—each with 16 layers and a hidden dimension of 1024, resulting in 0.6 billion parameters. 
The denoising multi-layer perceptron (MLP) consists of three blocks, each with a dimension of 1280. 
For spatial modeling, we use the encoder-decoder architecture from MAR~\cite{li2025autoregressive}.  
Following \citet{lin2024open}, we leverage a pre-trained and frozen variational autoencoder (VAE) as an image encoder to achieve spatio-temporal compression of the video, achieving $4{\times}4$ compression in the temporal dimension and $8{\times}8$ compression in the spatial dimension.
During training, we apply the masking and diffusion schedulers from~\citet{nichol2021improved}, using a masking ratio ranging from 0.7 to 1.0. In the inference phase, the ratio is gradually decreased from 1.0 to 0 according to a cosine schedule \cite{chang2023muse}.

~\\
\noindent
\textbf{Implementation.}
The captions and video clips are first tokenized into text tokens and visual tokens using a pre-trained language model~\cite{javaheripi2023phi} and an image encoder, respectively. These tokenized elements are then sequentially arranged into an input sequence that preserves their interleaved structure.
For example, the sequence is structured as follows: [\emph{caption$_{indiv\#1}$, clip$_{\#1}$, caption$_{indiv\#2}$, caption$_{joint\#1}$, clip$_{\#2}$ ...}], and so on.
Supervision is applied exclusively to the visual tokens through a diffusion loss~\cite{li2025autoregressive}.
For optimization, we use the AdamW optimizer~\cite{loshchilov2017decoupled} with $\beta_1 = 0.9$ and $\beta_2 = 0.95$, a weight decay of 0.02, and a base learning rate of $1 \times 10^{-4}$ in all experiments. We initialize the model weights using the T2V model NOVA-0.6B~\cite{deng2024nova} to accelerate convergence. All experiments are conducted on NVIDIA A100 40GB GPUs 

\subsection{Experimental Results}
\noindent
\textbf{Text Prompt Generation.}
To support evaluation, we generated 1,000 text prompts using seed keywords from VBench \cite{huang2024vbench}. Each prompt consists of six interconnected scenes that collectively form a coherent and engaging narrative. An example is shown in Table \ref{tab:prompt-example}.

\begin{table}[t!]
\small
\centering
\renewcommand{\arraystretch}{1.3}
\begin{tabular}{ p{7.5cm}}
\toprule
\textbf{Scene} \textbf{Description} \\
\midrule
\textbf{Scene \#1:}  \emph{``A curious little girl wearing a bright yellow dress tiptoes into a lush botanical garden, wide-eyed as she takes in the vibrant flowers and towering trees surrounded by crystal-clear ponds."} \\
\textbf{Scene \#2:}   \emph{``She spots a giant butterfly with shimmering blue wings fluttering over a bed of purple orchids and begins to follow it, her footsteps light and careful."} \\
\textbf{Scene \#3:}   \emph{``The butterfly leads her to a magnificent greenhouse, its glass walls reflecting the green world outside. Inside, tropical plants with oversized leaves spiral toward the ceiling."} \\
\textbf{Scene \#4:}   \emph{``Suddenly, the girl comes across an ancient, worn bench beneath a sprawling tree. She settles down and notices a squirrel nibbling on a tiny nut, staring curiously at her."} \\
\textbf{Scene \#5:}    \emph{``After feeding the squirrel a crumb from her pocket, the girl notices brilliant golden rays of sunlight breaking through the glass ceiling, lighting up the garden like a magical wonderland."} \\
\textbf{Scene \#6:}    \emph{``The butterfly lands gently on her shoulder, and she laughs in delight as the camera pans out, showing her peacefully seated amidst the blooming paradise."} \\
\bottomrule
\end{tabular}
\caption{An example prompt used for evaluation.}
\label{tab:prompt-example}
\end{table}

\subsubsection{Qualitative Results}
A video generation model trained on the large-scale Emu3 dataset serves as the baseline and is further fine-tuned on CI-VID.
In Figure~\ref{fig:qualitative}, we present a qualitative comparison between samples fine-tuned on CI-VID and those generated without finetuning, to demonstrate the capabilities acquired by the model after training on CI-VID. These samples are selected from the constructed text prompts. Notably, sample (a) is generated based on the prompt shown in Table~\ref{tab:prompt-example}.

As shown, the fine-tuned model exhibits improved stylistic and character consistency across the entire video sequence. It maintains a high level of uniformity in color, texture, and structure, effectively preserving character identity and environmental coherence across different clips. In contrast, the non-fine-tuned model fails to capture such cross-clip continuity. In contrast, while the base model can generate each scene based on the given prompt, it fails to establish meaningful relationships across scenes and may produce errors due to blindness to prior contextual information.
Moreover, the fine-tuned model not only adheres more faithfully to the input text instructions but also achieves natural and coherent camera transitions between clips, resulting in video sequences with enhanced narrative flow and storytelling quality.

\begin{table}[t!]
\centering
\small
\begin{tabular}{l|ccc}
\toprule
Metric & Win & Tie & Loss \\
\midrule
consistency & \textbf{90.0\%} & 6.5\% & 3.6\% \\
narrativity & \textbf{80.9\%} & 15.0\% & 4.1\% \\
correctness & \textbf{78.3\%} & 9.8\% & 11.9\% \\

\bottomrule
\end{tabular}
\caption{Human evaluation results in Win/Tie/Loss percentages comparing the fine‑tuned model against the base model.}
\label{Human Evaluation}
\label{tabel1}
\end{table}

\begin{table}[t!]
\centering
\resizebox{1.0\linewidth}{!}{
\setlength{\tabcolsep}{2pt}  
\begin{tabular}{l|cccc|cc}
\toprule
\footnotesize
\makecell{Model/\\Dimension} & 
\footnotesize
\makecell{Stylistic\\Consistency} & 
\footnotesize
\makecell{Entity\\Consistency} & 
\footnotesize
\makecell{Background\\Consistency} & 
\footnotesize
\makecell{Perspective\\Transition} & 
\footnotesize
\makecell{Text Prompt\\Alignment} & 
\footnotesize
\makecell{Visual\\Plausibility} \\
\midrule
\normalsize
\footnotesize Baseline & \normalsize 2.93 & \normalsize 2.84 & \normalsize 2.80 & \normalsize 3.02 & \normalsize 3.99 & \normalsize 3.25 \\
\footnotesize +CI-VID & \normalsize \textbf{3.83} & \normalsize \textbf{3.73} & \normalsize \textbf{3.75} & \normalsize \textbf{3.81} & \normalsize \textbf{4.07} & \normalsize \textbf{3.62} \\
\bottomrule
\end{tabular}
}
\caption{VLM-based evaluation results.}
\label{vlm_result}
\end{table}


\subsubsection{Quantitative Evaluation and Results}
\noindent
\textbf{Human Evaluation.}  
We conduct human evaluation for all comparison results. Each model output is represented as a row of merged keyframes—one selected from each video clip—as illustrated in Figure~\ref{fig:qualitative}. To avoid bias, model identities are anonymized and the top-bottom ordering is randomized.
Three full-time evaluators are tasked with comparing the outputs of two models side-by-side across three aspects:  
\emph{Consistency} (in terms of object, background, and visual style),  
\emph{Narrativity} (the coherence and storytelling quality of the clip sequence), and  
\emph{Factual Correctness} (faithfulness to the textual prompt, the correctness of visual content, and absence of visual distortions).  
Each comparison is labeled as either a \text{win}, \text{tie}, or \text{loss}. Evaluator agreement on consistency reaches 91\% (with ties) and 97\% (without ties). Final results are aggregated across all evaluators’ judgments and reported in Table~\ref{Human Evaluation}.
As shown in the example and summarized results, the model fine-tuned on CI-VID significantly outperforms the base model in three evaluated aspects.

~\\
\noindent
\textbf{VLM-based Evaluation.}
Following VBench \cite{huang2024vbench}, we employ VLMs to evaluate generated video sequences along six dimensions (shown in Table \ref{vlm_result}).
Specifically, Qwen2.5-VL-72B-Instruct \cite{bai2025qwen2} is asked to assign scores from 0 to 5 (very poor, poor, fair, good, excellent, very excellent) for each dimension based on the given prompt and video frames. 
For each sample, we evaluate both the full video sequence and pairwise clips. Final scores are the avg over six times evaluations (1 full + 5 pairwise). The VLM is calibrated with a reference example for scoring consistency.
These evaluation results are shown in Table \ref{vlm_result}.
The results show that CI-VID model \text{clearly outperforms} the baseline model on dimensions 1–4 (interleave related), and slightly surpasses that on the 6th. On the 5th dimension, it performs similar to the baseline model.

\begin{figure}[t!]
  \centering
  \includegraphics[page=1, width=1.0\linewidth]{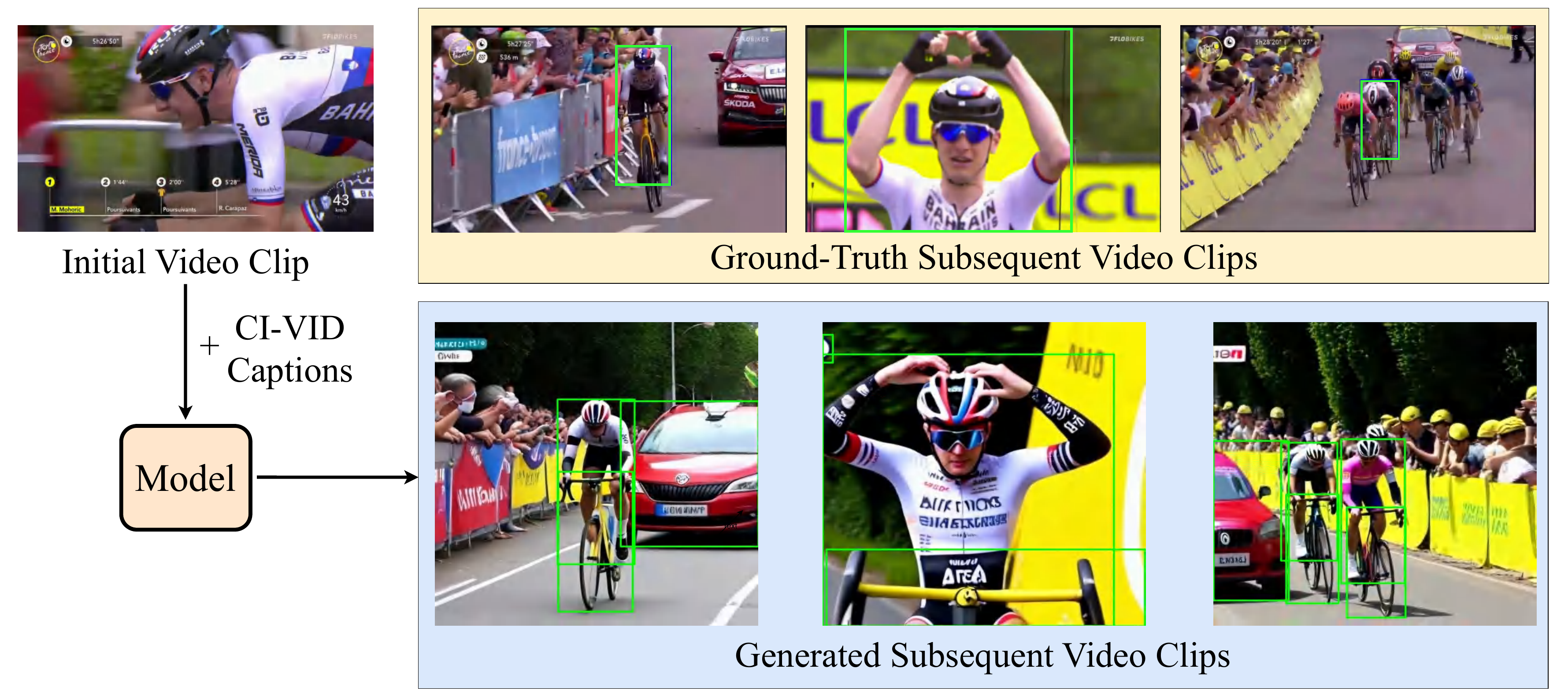}
  \caption{Example of similarity-based evaluation setup.}
  \label{fig:Similarity}
\end{figure}

\begin{table}[t!]
\footnotesize
\centering
\resizebox{0.98\linewidth}{!}{
\renewcommand{\arraystretch}{1.0}  
\begin{tabular}{l|ccc|ccc}
\toprule
Model/ & \multicolumn{3}{c|}{Whole} & \multicolumn{3}{c}{Object} \\
\cmidrule(lr){2-4} \cmidrule(lr){5-7}
Metric  & CLIP & 1 - LPIPS & SSIM       & CLIP & 1 - LPIPS & SSIM \\
\midrule
Baseline   & 0.512 & 0.309 & 0.199 & 0.601 & 0.360 & 0.278 \\
+CI-VID   & \textbf{0.670} & \textbf{0.381} & \textbf{0.272} & \textbf{0.702} & \textbf{0.412} & 0.\textbf{391} \\
\bottomrule
\end{tabular}
}
\caption{Similarity evaluation results. Metrics include CLIP similarity, inverse LPIPS, and SSIM. Higher is better ($\uparrow$).}
\label{sim_re}
\end{table}

~\\
\noindent
\textbf{Similarity-based Evaluation.}
We construct a similarity-based evaluation dataset based on CI-VID. To prevent data leakage, all test samples and their source-related counterparts are excluded from the training set.
As shown in Figure~\ref{fig:Similarity}, given an initial video clip, the model is tasked with generating subsequent clips. Evaluation is performed by measuring both overall and object-level similarity between the generated and ground-truth videos.
We employ YOLO-World-L \cite{cheng2024yolo} as object detector to identify key objects, and manually retain narrative-relevant entities as ground-truth references. Each ground-truth clip consists of three reference frames.
We adopt three widely used similarity metrics: CLIP similarity \cite{radford2021learning}, which captures semantic alignment between frames and text; 1–LPIPS, which measures perceptual closeness; and SSIM, which assesses structural similarity. 
For CLIP similarity, we use the ViT-H/14 variant pretrained on the LAION-2B \cite{schuhmann2022laion}, and compute the cosine similarity.
As shown in Table~\ref{sim_re}, our model achieves superior performance across all three metrics, at both the holistic and object-specific levels.

\section{Conclusion}
We introduce CI-VID, a dataset that moves beyond isolated text-to-video (T2V) generation toward text-and-video-to-video (T\&V2V) generation, enabling models to produce coherent multi-scene video sequences.
In addition, we design a multi-dimensional benchmark to evaluate the task of coherent video sequence generation from both human and automatic perspectives. 
Experimental results on this benchmark further validate the effectiveness and utility of the proposed CI-VID.

{
    \small
    \bibliographystyle{ieeenat_fullname}
    \bibliography{main}
}


\appendix
\section{Experiment Details}

The diffusion loss used in our experiment is formulated as:
\[
\mathcal{L}({x}_{n}\,|\,{z}_{n}) = \mathbb{E}_{\varepsilon, t}\left[\left\|\epsilon - {\epsilon}_{\theta}\left({x}_{n}^{t} \mid t, {z}_{n}\right)\right\|^{2}\right],
\]
where $\epsilon$ is a Gaussian vector sampled from $\mathcal{N}(\mathbf{0}, \mathbf{I})$. The noisy input ${x}_{n}^{t}$ is generated from the original sample ${x}_{n}$ as
\[
x_{n}^{t} = \sqrt{\bar{\alpha}_{t}} x_{n} + \sqrt{1 - \bar{\alpha}_{t}} \epsilon,
\]
where $\bar{\alpha}_{t}$ denotes a noise schedule indexed by time step $t$. The noise estimator $\epsilon_{\theta}$, parameterized by $\theta$ and implemented as a stack of MLP blocks, takes ${x}_{n}^{t}$ as input and is conditioned on both $t$ and ${z}_{n}$. 

We sample $t$ four times during each training iteration for every image. During inference, we initialize $x_n^T$ with noise sampled from $\mathcal{N}(\mathbf{0}, \mathbf{I})$, and progressively denoise it to $x_n^0$ using the following sequential steps:
\[
x_{n}^{t-1} = \frac{1}{\sqrt{\alpha_{t}}}\left(x_{n}^{t} - \frac{1-\alpha_{t}}{\sqrt{1-\bar{\alpha}_{t}}} \epsilon_{\theta}(x_{n}^{t} \mid t, z_{n})\right) + \sigma_{t} \epsilon,
\]
where $\sigma_{t}$ denotes the noise level at time step $t$, and $\epsilon$ is again drawn from $\mathcal{N}(\mathbf{0}, \mathbf{I})$.





\section{CI-VID Examples}
Several video sequences are extracted from CI-VID and presented below to illustrate the dataset's characteristics. Additional examples with captions can be found in our GitHub repository.

\begin{figure}[t!]
  \centering

  \begin{subfigure}{\linewidth}
    \centering
    \includegraphics[width=0.95\linewidth]{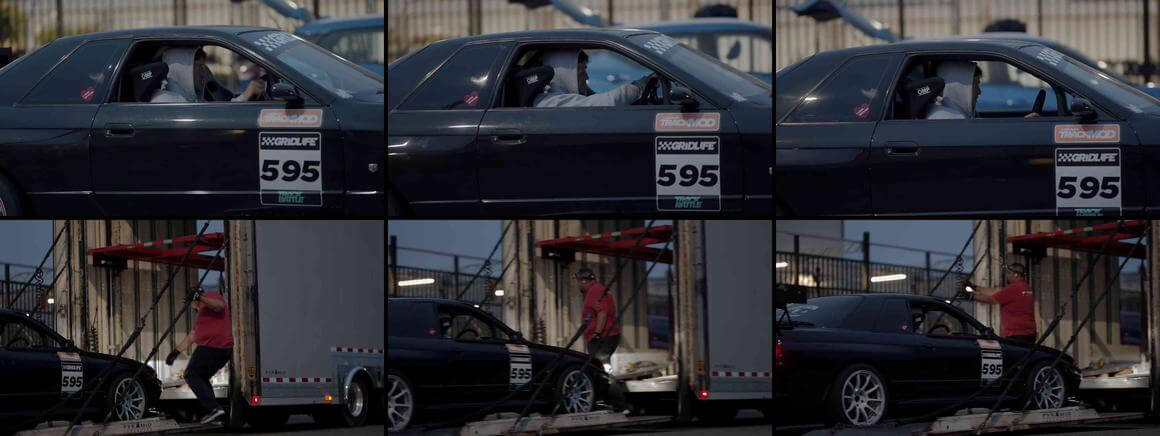}
    \caption{ \href{https://flagchat.ks3-cn-beijing.ksyuncs.com/ymju/CI_VID_captions/003_caption.txt}{download corresponding captions}}
  \end{subfigure}

  \vspace{4mm} 

  \begin{subfigure}{\linewidth}
    \centering
    \includegraphics[width=0.95\linewidth]{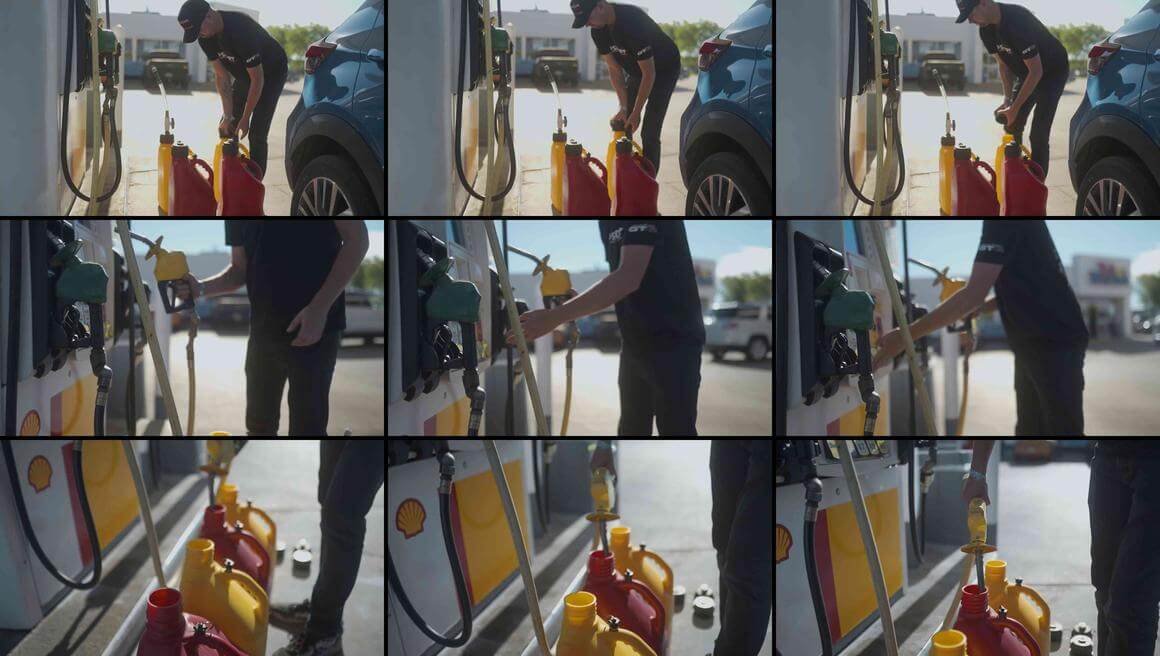}
    \caption{\href{https://flagchat.ks3-cn-beijing.ksyuncs.com/ymju/CI_VID_captions/006_caption.txt}{download corresponding captions}}
  \end{subfigure}

  \vspace{4mm}

  \begin{subfigure}{\linewidth}
    \centering
    \includegraphics[width=0.95\linewidth]{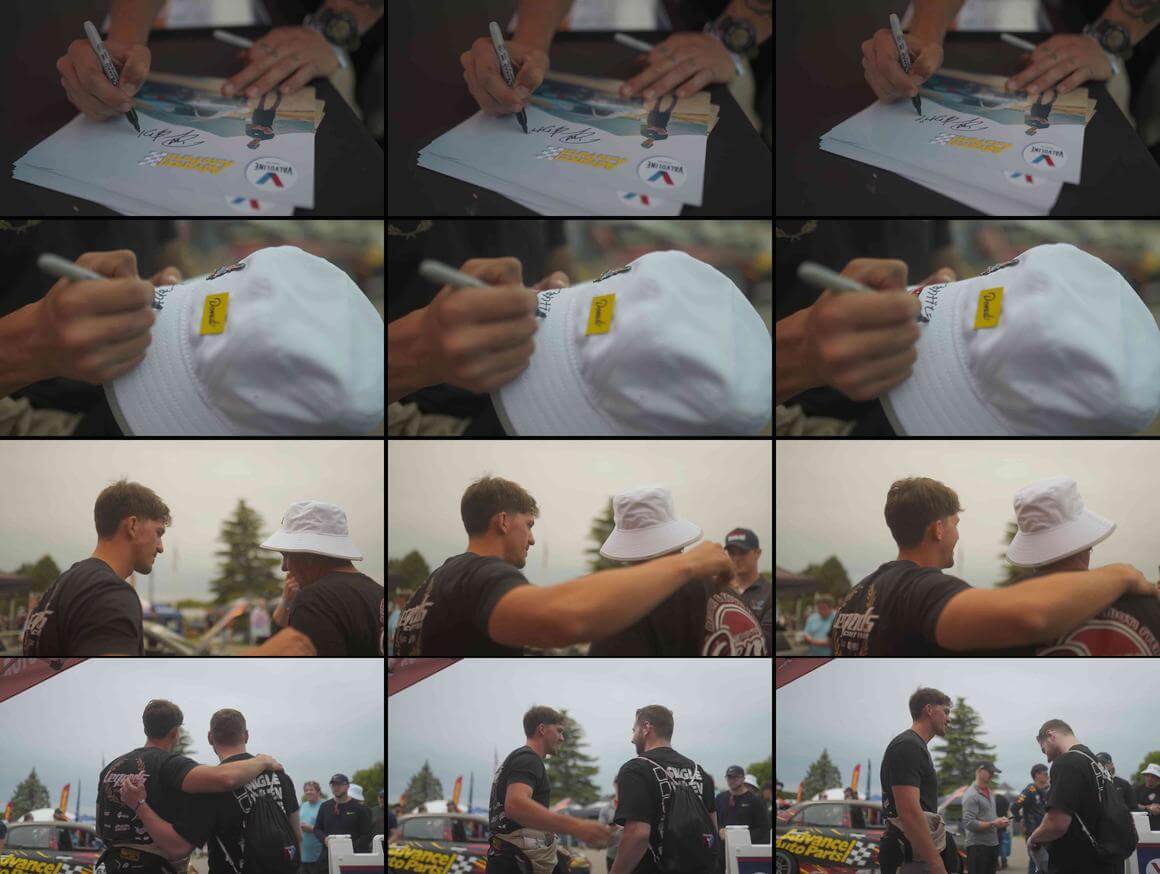}
    \caption{\href{https://flagchat.ks3-cn-beijing.ksyuncs.com/ymju/CI_VID_captions/001_caption.txt}{download corresponding captions}}
  \end{subfigure}
  
  \caption{CI-VID Examples (1/6): Video sequences are extracted from the same original video. Each row corresponds to one video clip.}

\end{figure}

\begin{figure}[t!]
  \centering

  \begin{subfigure}{\linewidth}
    \centering
    \includegraphics[width=0.95\linewidth]{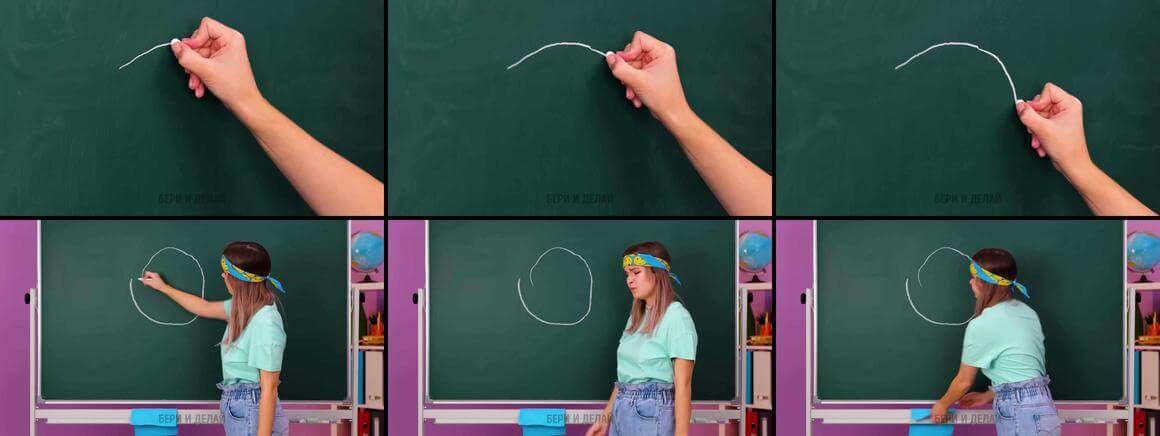}
    \caption{\href{https://flagchat.ks3-cn-beijing.ksyuncs.com/ymju/CI_VID_captions/020_caption.txt}{download corresponding captions}}
  \end{subfigure}

  \vspace{4mm} 

  \begin{subfigure}{\linewidth}
    \centering
    \includegraphics[width=0.95\linewidth]{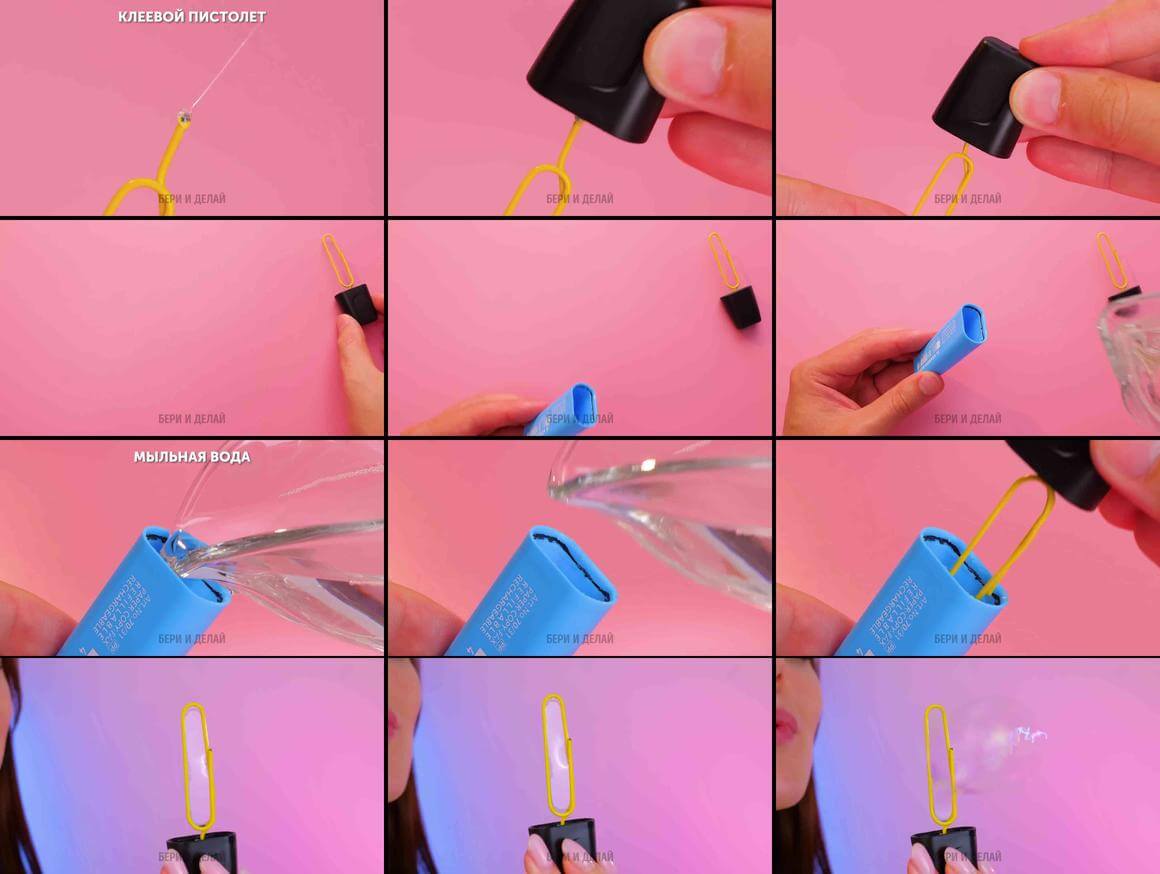}
    \caption{\href{https://flagchat.ks3-cn-beijing.ksyuncs.com/ymju/CI_VID_captions/017_caption.txt}{download corresponding captions}}
  \end{subfigure}

  \vspace{4mm}

  \begin{subfigure}{\linewidth}
    \centering
    \includegraphics[width=0.95\linewidth]{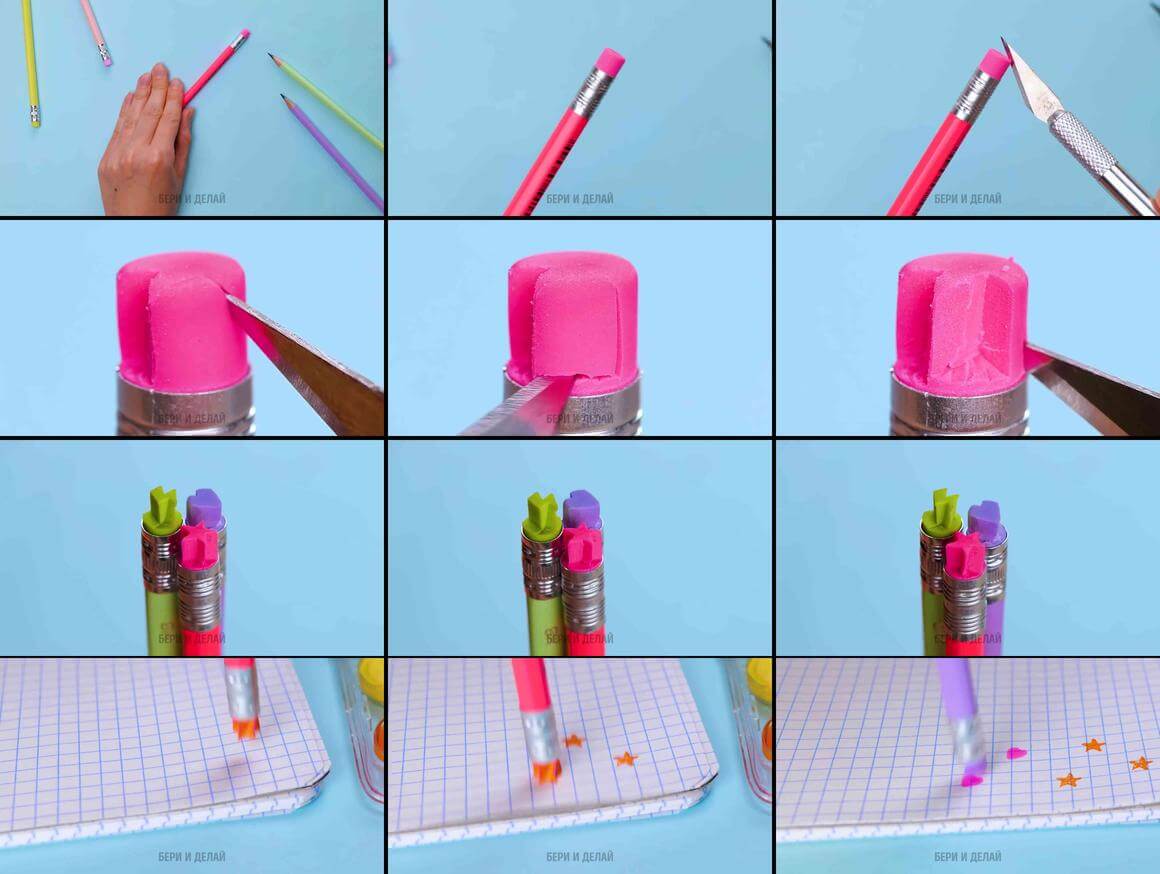}
    \caption{\href{https://flagchat.ks3-cn-beijing.ksyuncs.com/ymju/CI_VID_captions/023_caption.txt}{download corresponding captions}}
  \end{subfigure}
  
  \caption{CI-VID Examples (2/6): Video sequences are extracted from the same original video. Each row corresponds to one video clip.}  
\end{figure}

\begin{figure}[t!]
  \centering

  \begin{subfigure}{\linewidth}
    \centering
    \includegraphics[width=0.95\linewidth]{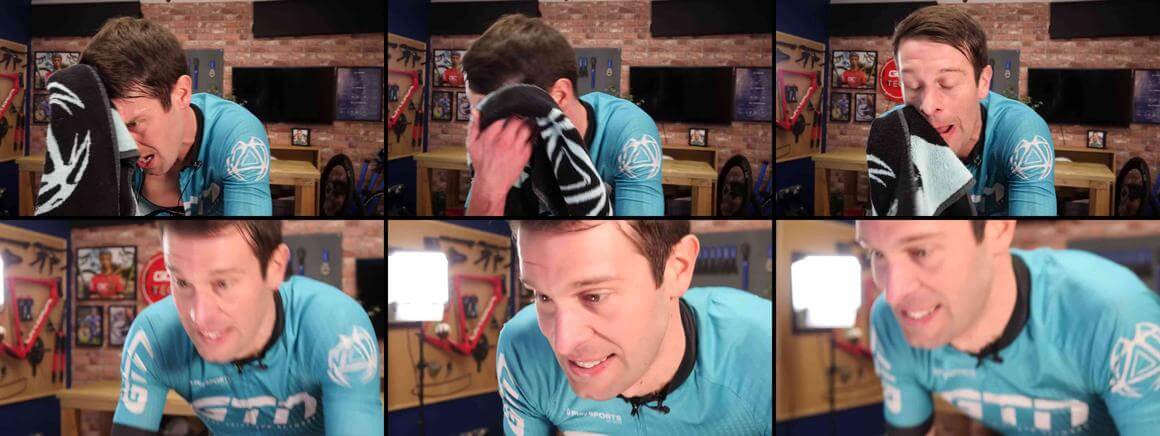}
    \caption{\href{https://flagchat.ks3-cn-beijing.ksyuncs.com/ymju/CI_VID_captions/064_caption.txt}{download corresponding captions}}
  \end{subfigure}

  \vspace{4mm} 

  \begin{subfigure}{\linewidth}
    \centering
    \includegraphics[width=0.95\linewidth]{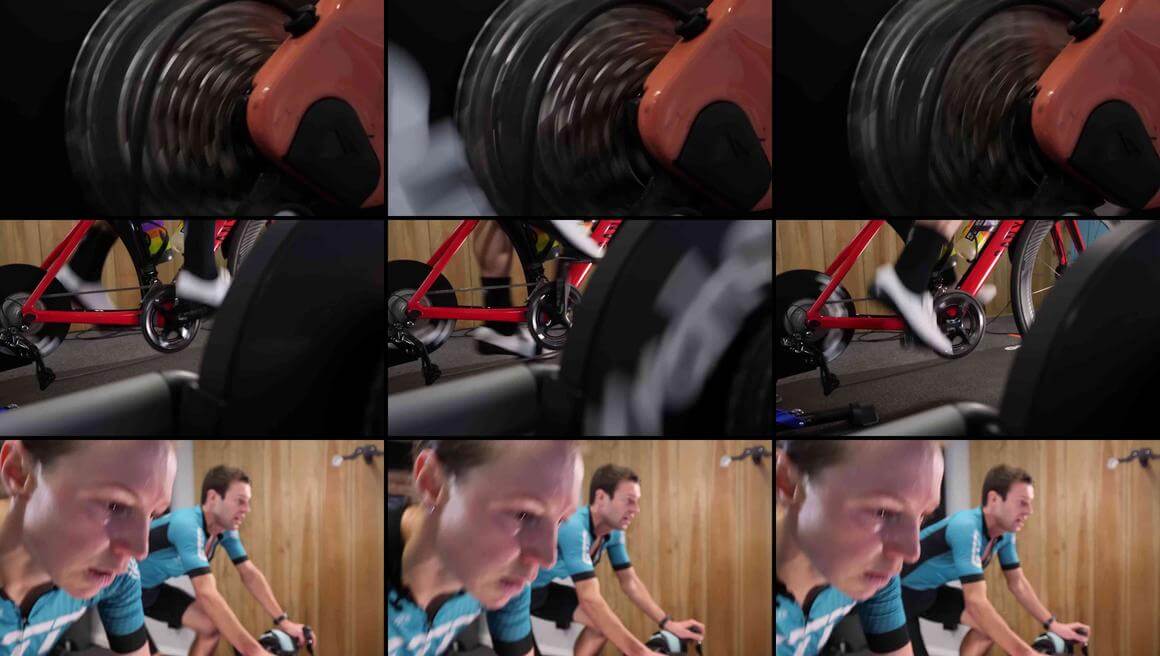}
    \caption{\href{https://flagchat.ks3-cn-beijing.ksyuncs.com/ymju/CI_VID_captions/067_caption.txt}{download corresponding captions}}
  \end{subfigure}

  \caption{CI-VID Examples (3/6): Video sequences are extracted from the same original video. Each row corresponds to one video clip.}    
  
\end{figure}

\begin{figure}[t!]
  \centering

  \begin{subfigure}{\linewidth}
    \centering
    \includegraphics[width=0.95\linewidth]{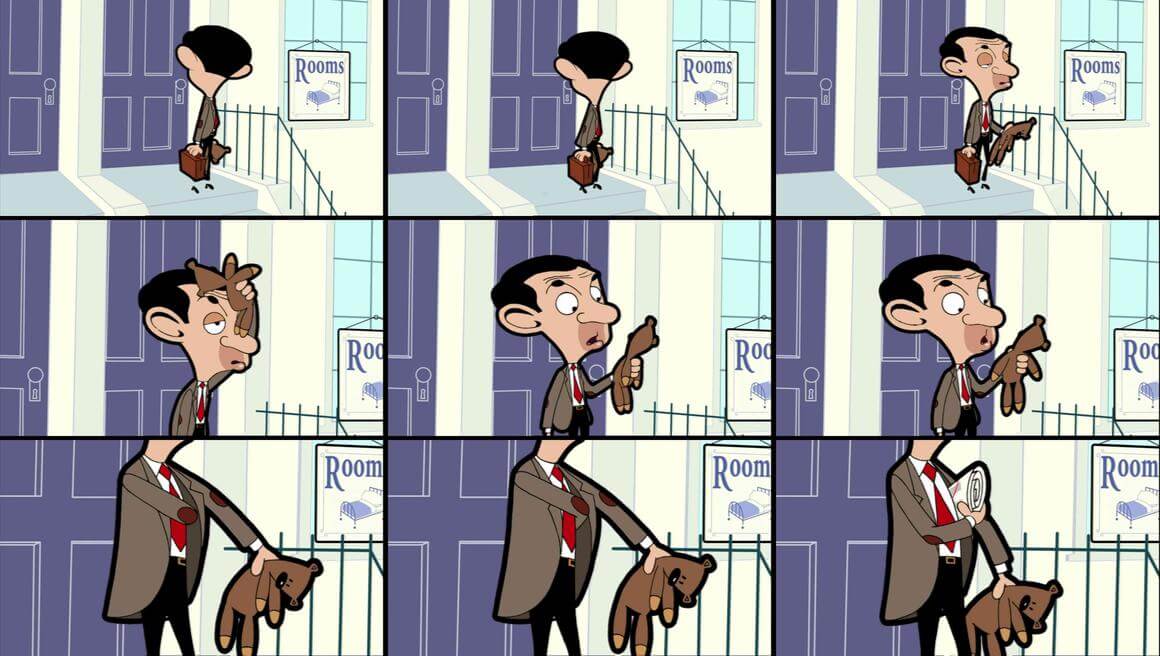}
    \caption{\href{https://flagchat.ks3-cn-beijing.ksyuncs.com/ymju/CI_VID_captions/109_caption.txt}{download corresponding captions}}
  \end{subfigure}

  \vspace{4mm} 

  \begin{subfigure}{\linewidth}
    \centering
    \includegraphics[width=0.95\linewidth]{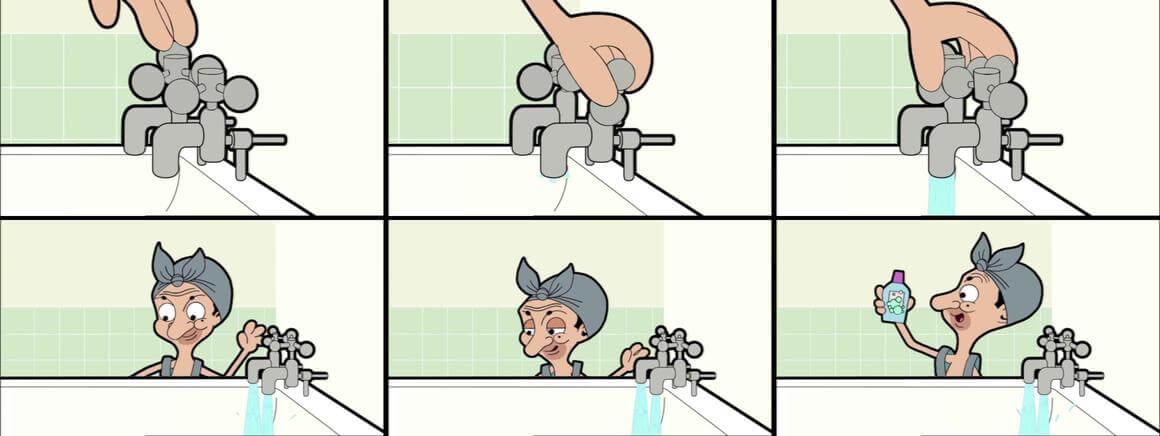}
    \caption{\href{https://flagchat.ks3-cn-beijing.ksyuncs.com/ymju/CI_VID_captions/116_caption.txt}{download corresponding captions}}
  \end{subfigure}

  \caption{CI-VID Examples (4/6): Video sequences are extracted from the same original video. Each row corresponds to one video clip.}    
  
\end{figure}

\begin{figure}[t!]
  \centering

  \begin{subfigure}{\linewidth}
    \centering
    \includegraphics[width=0.95\linewidth]{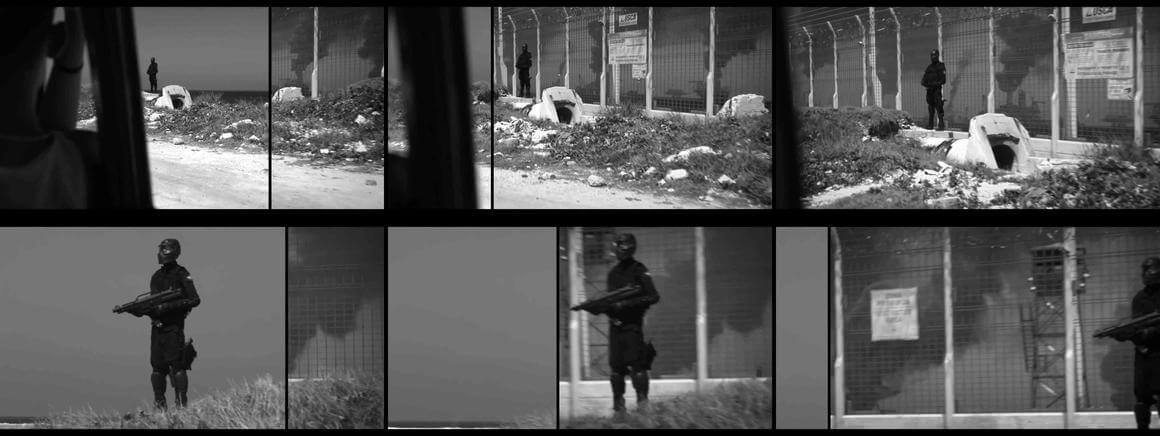}
    \caption{\href{https://flagchat.ks3-cn-beijing.ksyuncs.com/ymju/CI_VID_captions/076_caption.txt}{download corresponding captions}}
  \end{subfigure}

  \vspace{4mm} 

  \begin{subfigure}{\linewidth}
    \centering
    \includegraphics[width=0.95\linewidth]{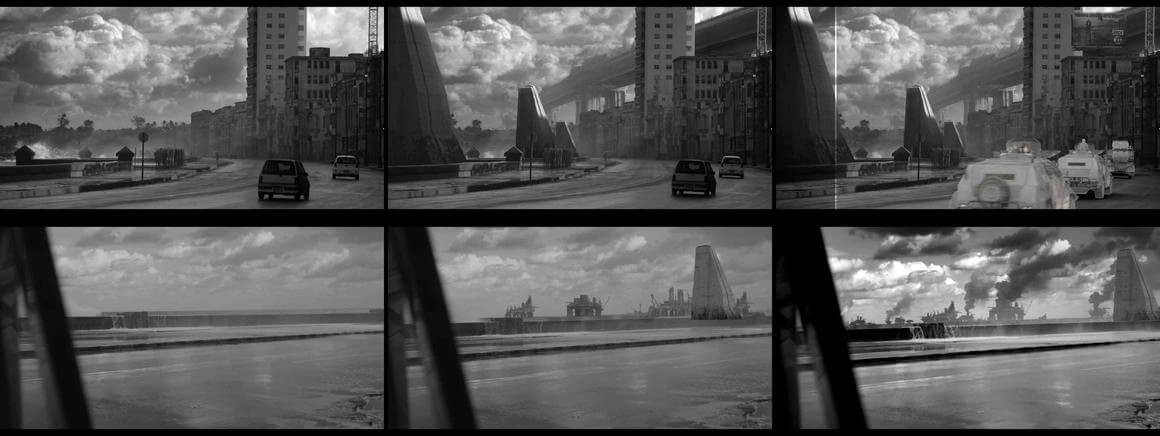}
    \caption{\href{https://flagchat.ks3-cn-beijing.ksyuncs.com/ymju/CI_VID_captions/077_caption.txt}{download corresponding captions}}
  \end{subfigure}

  \caption{CI-VID Examples (5/6): Video sequences are extracted from the same original video. Each row corresponds to one video clip.}    
  
\end{figure}

\begin{figure}[t!]
  \centering

  \begin{subfigure}{\linewidth}
    \centering
    \includegraphics[width=0.95\linewidth]{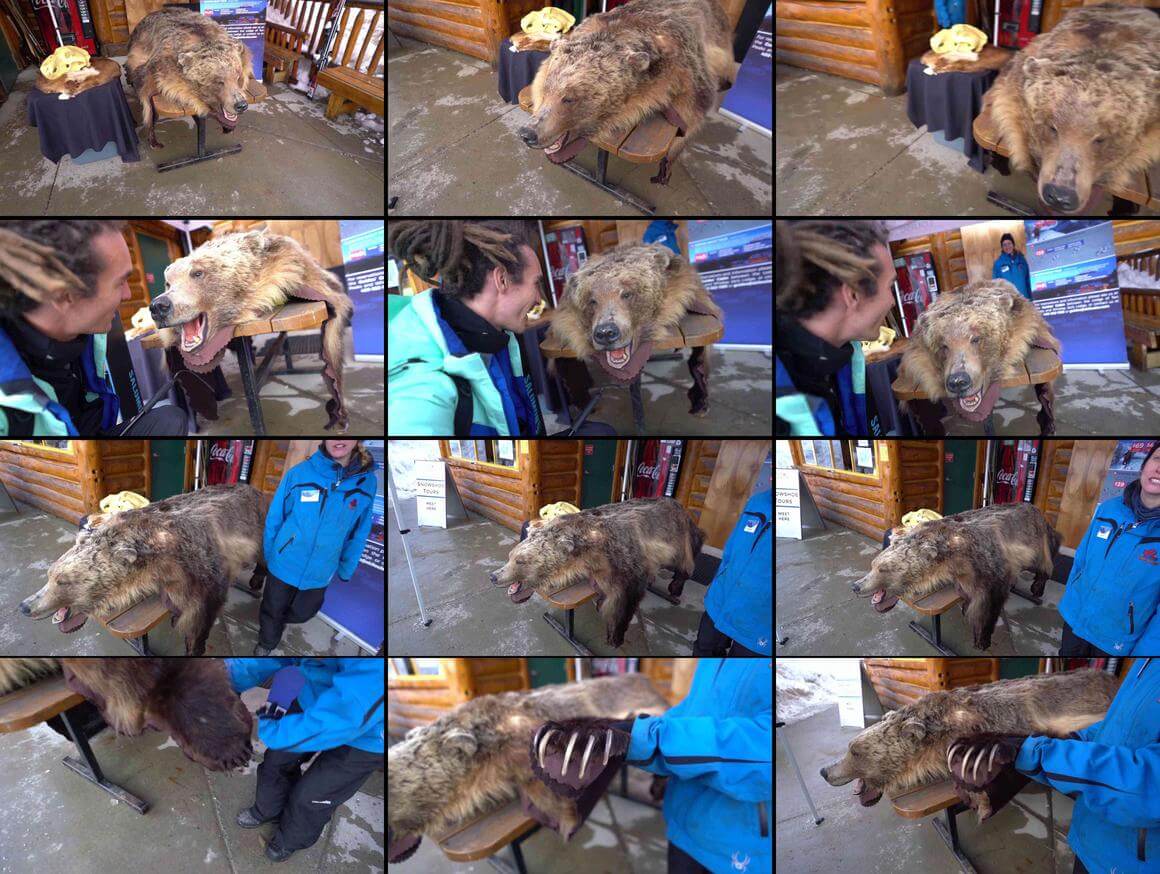}
    \caption{\href{https://flagchat.ks3-cn-beijing.ksyuncs.com/ymju/CI_VID_captions/096_caption.txt}{download corresponding captions}}
  \end{subfigure}

  \caption{CI-VID Examples (6/6): Video sequences are extracted from the same original video. Each row corresponds to one video clip.}
  
\end{figure}

\end{document}